\documentclass[conference]{IEEEtran}
\IEEEoverridecommandlockouts
\usepackage{cite}
\usepackage{amsmath,amssymb,amsfonts}
\usepackage{algorithmic}
\usepackage{graphicx}
\usepackage{textcomp}
\usepackage{soul}
\usepackage{xcolor}         
\usepackage{comment}
\usepackage[bookmarks=false]{hyperref}
\usepackage{hypcap}

\def\BibTeX{{\rm B\kern-.05em{\sc i\kern-.025em b}\kern-.08em
    T\kern-.1667em\lower.7ex\hbox{E}\kern-.125emX}}
\begin{document}
\bstctlcite{IEEEexample:BSTcontrol} 
\title{Integrating Ontology Design with the CRISP-DM in the context of Cyber-Physical Systems Maintenance}

\author{
    \IEEEauthorblockN{
        Milapji Singh Gill\IEEEauthorrefmark{1},
        Tom Westermann\IEEEauthorrefmark{1},
        Gernot Steindl\IEEEauthorrefmark{2},
        Felix Gehlhoff\IEEEauthorrefmark{1}, Alexander Fay\IEEEauthorrefmark{3}%
    } 
    
    \IEEEauthorblockA{
        \IEEEauthorrefmark{1}Institute of Automation Technology\\
        \textit{Helmut Schmidt University Hamburg, Germany}\\
        {\tt\small \{milapji.gill, tom.westermann, felix.gehlhoff\}@hsu-hh.de}\\
        \IEEEauthorrefmark{2}Institute of Computer Engineering\\
        \textit{TU Wien, Austria}\\
        {\tt\small \ gernot.steindl@tuwien.ac.at}\\
        \IEEEauthorrefmark{3}Chair of Automation\\
        \textit{Ruhr University Bochum, Germany}\\
        {\tt\small \ alexander.fay@rub.de}\\
       }
       \vspace{-1cm}
}
\maketitle

\begin{abstract}
In the following contribution, a method is introduced that integrates domain expert-centric ontology design with the Cross-Industry Standard Process for Data Mining (CRISP-DM). This approach aims to efficiently build an application-specific ontology tailored to the corrective maintenance of Cyber-Physical Systems (CPS). The proposed method is divided into three phases. In phase one, ontology requirements are systematically specified, defining the relevant knowledge scope. Accordingly, CPS life cycle data is contextualized in phase two using domain-specific ontological artifacts. This formalized domain knowledge is then utilized in the CRISP-DM to efficiently extract new insights from the data. Finally, the newly developed data-driven model is employed to populate and expand the ontology. Thus, information extracted from this model is semantically annotated and aligned with the existing ontology in phase three. The applicability of this method has been evaluated in an anomaly detection case study for a modular process plant.
\end{abstract}


\section{Introduction}\label{introduction}

Downtime due to faults occurring during operation significantly reduces the profitability of using Cyber-Physical Systems (CPS) \cite{Gill.2023b}. To mitigate these impacts, efficient and effective corrective maintenance is essential. However, the growing complexity of CPS, driven by sophisticated technology integration and increased connectivity, complicates the process of fault localization \cite{Westermann.2023}. Consequently, technicians need assistance systems that deliver accurate information, enabling swift restoration of system operations\cite{Steindl.2021, Gill.2023b}. The use of Artificial Intelligence (AI) to automate activities such as anomaly detection and fault diagnosis proves beneficial in realizing valuable digital services to accelerate the maintenance process \cite{Steindl.2021, Zhou.2019}. 
In order to generate valuable insights with data-driven models, a substantial volume of operational and maintenance data is required. Yet, this alone is often insufficient. Value-adding digital maintenance services need integration with other information and prior knowledge \cite{Zhou.2021, Rueden.2021}. This is even more important as the lack of domain knowledge frequently impedes the work of data experts \cite{Zhang.2018, Dou.2015}. Additionally, CPS information from digital models, generated during the engineering phase, is highly relevant in several maintenance activities\cite{Hildebrandt.2020, Gill.2023b}. As a result, effective and efficient corrective maintenance of CPS requires life-cycle-wide management and combination of data, information, and knowledge. Given these prerequisites, integrating data from heterogeneous sources poses a significant challenge \cite{Hildebrandt.2020}. 

Using the Digital Twin concept along with domain-specific ontologies effectively consolidates information for digital maintenance services\cite{Gill.2022b, Reinpold.2024}. On the one hand, the introduction of an ontology improves the integration and accessibility of data and digital models across the CPS life cycle \cite{Hildebrandt.2020, Steindl.2021}. Consequently, domain knowledge can be leveraged for data analytics to efficiently generate new insights \cite{Dou.2015, Rueden.2021}. On the other hand, due to their expandability and adaptability, existing ontologies can be efficiently populated and enriched with collected data and newly developed data-driven models \cite{Sahlab.2021, Steindl.2021}. This includes the combination of extracted information with insights from the operational and maintenance phases with prior knowledge from engineering \cite{Gill.2023b, Zhou.2021}. Therefore, ontology design can benefit from data-driven approaches to gradually enhance the existing ontology with new knowledge. However, building application-specific ontologies involves substantial effort, as each maintenance use case requires unique information and data-driven models. The described synergies between ontologies and data-driven approaches are rarely exploited systematically. Hence, the goal of this contribution is to introduce a method that efficiently combines ontology design with data-driven modeling process steps. Firstly, this approach aims to effectively utilize existing domain knowledge captured in an ontology for the development of new data-driven models. Secondly, it employs these new data insights to expand the existing ontology for the maintenance application.

This paper is structured as follows: Related works are analyzed in Sec. \ref{Related Work}, followed by the specification of relevant requirements in Sec. \ref{Requirements}. The method for building an application-specific ontology in the context of CPS maintenance is detailed in Sec. \ref{contribution}. The applicability of this method is evaluated in an anomaly detection case study for a modular process plant in Sec. \ref{Evaluation} and discussed in Sec. \ref{Discussion}. The paper concludes with a summary and an outlook in Sec. \ref{Summary and Outlook}.

\section{Related Work}\label{Related Work}
In the following section, related works are presented. Initially, focus areas in ontology design are analyzed. Subsequently, possibilities for integration with data-driven models regarding the corrective maintenance of CPS are examined.

Given the significant effort required to build an application-specific ontology and the scarcity of ontology experts, the approach by Hildebrandt et al. \cite{Hildebrandt.2020} is particularly valuable. The authors advocate for creating reusable, modular ontological artifacts that incorporate domain-specific resources such as norms and standards. Ontological artifacts, such as Ontology Design Patterns (ODPs), include essential elements for ontology development and are created using semantic web technologies. The method starts with a systematic specification of ontology requirements through Competency Questions (CQs), conducted by software and domain experts. This step is followed by the creation of a Lightweight Ontology (LWO), represented as a Unified Modeling Language (UML) class diagram. Existing LWO Design Patterns are either utilized or new ones are developed and aligned for this purpose. Subsequently, an ontology expert transforms this LWO into a Heavyweight Ontology (HWO), which includes a terminological box (T-Box) in Web Ontology Language (OWL) format. For a maintenance use case, the assertional Box (A-Box) is populated with both static and dynamic data. This method is particularly suited for CPS ontology development, with information on function, structure, and behavior extracted from engineering artifacts being especially important. Poveda-Villalon et al.'s Linked Open Terms (LOT) method \cite{PovedaVillalon.2022}, while not specifically addressing CPS, further emphasizes the maintenance of ontologies in response to new insights and knowledge. 

Further research highlights the distinct advantage of combining ontologies and data-driven models \cite{Steindl.2019, VogelHeuser.2021,Dou.2015}. This is especially beneficial for CPS maintenance, as the amount of data increases throughout its life cycle, while the number of knowledge-driven digital models decreases \cite{VogelHeuser.2021}. A major area of research focuses on integrating domain knowledge into the data modeling process in order to facilitate data integration and understanding. Key terms associated with this include Semantic Data Mining and Informed Machine Learning\cite{Rueden.2021, Dou.2015}. Additionally, there are efforts to populate and enrich existing ontologies using data-driven models. Ontology population refers to using data-driven models to efficiently instantiate the A-Box \cite{VogelHeuser.2021}. Conversely, ontology enrichment primarily involves expanding the T-Box with additional concepts and relations that have been identified through data-driven models \cite{Westermann.2023}. Such combining approaches have already been applied in the context of corrective maintenance \cite{Zhou.2021, Steindl.2021, Gill.2023b}. For example, Zhou et al. \cite{Zhou.2019} use a Gaussian mixture density hidden Markov model (CGHMM) for vibration analysis of rolling bearings, integrating the results with an ontology to enhance fault diagnosis. Similarly, Steindl et al. \cite{Steindl.2021} examine a combination approach for a thermal heating process. An ontology defines behavior and structure to contextualize operational data, which then feeds into a linear autoregressive with exogenous input (ARX) model. Identified anomalies are accordingly used to expand the ontology.

Given the diversity of data-driven models (e.g. classification or regression) employed for every use case, a methodological framework is necessary. The Cross-Industry Standard Process for Data Mining (CRISP-DM), a de facto standard for data-driven projects, integrates Data Mining (DM) and Machine Learning (ML) techniques to derive insights from data \cite{Wirth., Huber.2019}. This process model, which comprises steps like business understanding, data understanding, data preparation, modeling, evaluation, and deployment, is widely applied \cite{Huber.2019, Gill.2023a}. Thus, this process model is subsequently used in combination with ontology design. The authors of this contribution have previously proposed a method that incorporates ontological artifacts into the CRISP-DM, with a focus limited to the modeling step \cite{Gill.2023a}. The analysis of using ontologies at the modeling step to increase efficiency was not extensively covered, considering the promising advancements in automated ML (AutoML) solutions \cite{He.2021}. However, a complete combination of CRISP-DM with ontology design, including evaluation and deployment, has not yet been examined in the context of CPS.

\section{Requirements}\label{Requirements}

To design a method for developing and expanding ontologies with data-driven models for CPS maintenance, requirements (\textbf{R}) are established, based on the insights from Sec.\ref{Related Work}.

\textbf{R1: Systematic specification and documentation of ontology requirements} 

When developing and expanding ontologies, establishing systematic steps for gathering and documenting application-specific ontology requirements is essential \cite{Hildebrandt.2020, PovedaVillalon.2022, Gill.2023a}. This process aims to identify, at an early stage, the essential knowledge required for digital maintenance services by specifying CQs. These specifications are pertinent for subsequent method steps, enabling early monitoring and validation of created ontological artifacts. Additionally, this documentation can assist in future projects by simplifying the requirements comparison across different applications \cite{Hildebrandt.2020}.

\textbf{R2: Contextualization of CPS life cycle data} 

Efficiency losses are significant during data understanding and preparation steps, often due to the extensive time required for these activities \cite{VogelHeuser.2021, Dou.2015}. Research has demonstrated the immense value of incorporating CPS domain knowledge into these steps \cite{VogelHeuser.2021, Gill.2023a, Rueden.2021}. Beyond the issue of understanding data, relevant life cycle data is scattered across different silos, making the integration challenging \cite{Hildebrandt.2020, Gill.2023a}. Consequently, it's essential to systematically equip data experts with the information from a domain ontology, ensuring unified semantics \cite{Dou.2015, Steindl.2019}. 

\textbf{R3: Semantic annotation of extracted information} 

As described in the previous section, data-driven models, e.g. with new insights from operations and maintenance, are well-suited for efficiently populating (A-Box) or expanding (T-Box) an existing ontology. Data-driven models generated in the modeling step of the CRISP-DM can greatly differ based on the specific maintenance context \cite{Steindl.2021, Zhou.2019, Gill.2023b}. Given this variety, systematic steps are essential for expanding the existing ontology with newly generated data-driven models. This process involves semantically annotating the extracted information from the data-driven model, followed by systematically aligning any new concepts and relations with the existing ontology \cite{Zhou.2021, Dou.2015, VogelHeuser.2021}.

\textbf{R4: Modular, domain-specific ontological artifacts} 

Employing and documenting reusable modular ODPs aids in lessening the modeling efforts for application-specific ontologies with regard to repetitive tasks \cite{Hildebrandt.2020, Gill.2023a}. For any application-specific ontology, it is important to either align existing domain-specific ODPs within the development method or to create new ones as needed. These ODPs should adhere to standards, which promotes a consistent understanding of concepts and relationships within the ontology \cite{Hildebrandt.2020}. 

\textbf{R5: Domain expert-centric approach} 

Developing, expanding, and using an ontology for maintenance applications involves collaboration among diverse experts \cite{Hildebrandt.2020, Gill.2023a, PovedaVillalon.2022}. Each expert contributes unique knowledge (e.g. software, domain, data, ontologies), which must be systematically integrated into specific steps of the method. Involving every expert in all steps can lead to unnecessary costs and inefficiencies. Consequently, it is imperative to clearly define each expert's role, whether as a developer or consultant. Moreover, the method should adopt a domain expert-centric approach to address the scarcity of ontology experts and to ensure that relevant domain knowledge is formalized \cite{Hildebrandt.2020}. Ontological artifacts, therefore, should be comprehensible to experts (e.g. LWO) serving in development roles.

\section{Methodology}\label{contribution}

\subsection{Overview}
The method depicted in Fig.1 is designed to efficiently develop an application-specific ontology tailored to the corrective maintenance of CPS. Building on this foundation, digital services will be crafted to query, generate, and update information within the ontology. The steps are categorized into two swimlanes, which separate the CRISP-DM as well as domain expert-centric ontology design steps. The method is detailed through individual steps, highlighting the input and output artifacts, as well as the involvement of diverse experts in varied roles (such as developing and consulting).
Moreover, it is divided into three distinct phases. In the method's initial phase, a systematic specification of requirements for the development and expansion of the application-specific ontology is essential. The main objective is to formulate CQs drawn from the maintenance process and the digital services that need to be developed (see \textbf{R1}). The second phase aims to enhance the efficiency of extracting insights from data collected throughout the CPS life cycle. In this phase, a modular ontology is built to contextualize CPS data and incorporate domain knowledge for the subsequent data-driven modeling (see \textbf{R2}). In the third phase, extracted information is semantically annotated, followed by its alignment with the previously developed modular ontology. Subsequently, all created ontological artifacts must be made available for the deployment of digital maintenance services (see \textbf{R3}). In all phases, domain-specific, modular ontological artifacts are created or reused (see \textbf{R4}). These artifacts vary in their level of formality. Moreover, blue ones are pre-existing and documented from past projects, while green ones are newly designed for the specific project. Method steps range from manual to (semi-) automated, and fully automated steps. Key roles identified include domain, software, data and ontology experts (see \textbf{R5}). Domain experts, including technicians and engineers, possess knowledge in CPS maintenance. Software experts, such as architects and developers, manage software architecture and digital service implementation. Data experts, including data scientists, data analysts and data engineers, extract relevant insights from data. Ontology experts focus on semantic web technologies to guide ontology development.

\begin{figure*}[ht]
\centering
\includegraphics[width=1\textwidth, height= 9.4cm]{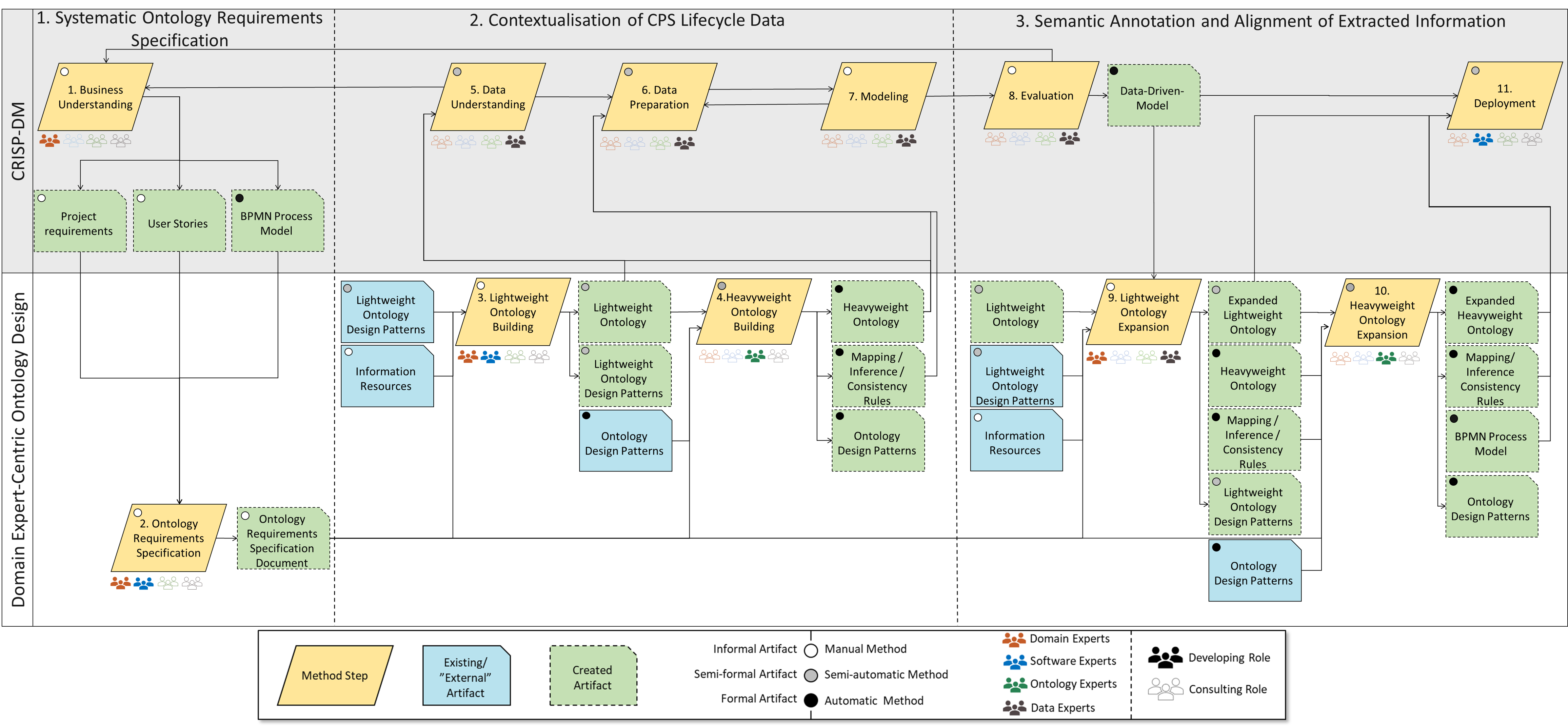}
\caption{Combination of domain expert-centric ontology design and the CRISP-DM }
\label{fig:Approach}
\end{figure*}

\subsection{Systematic Ontology Requirements Specification}
The goal of \textit{Step 1} (business understanding) is to derive the assistance needs for the specific CPS maintenance process. In this step, domain experts determine the information needs in the form of \textit{User Stories}. Standardized templates can be used for this purpose, which are textually filled out by the potential users of the assistance application. The project requirements define target metrics and criteria for evaluating interim results. Additionally, it's essential to model the examined maintenance process section to grasp the technician's journey through technical and administrative maintenance activities. The \textit{Business Process Modeling Notation (BPMN) }offers a standardized notation and the capability to generate an  Extensible Markup Language (XML) schema for process automation with a workflow engine \cite{Rosing.}. It serves as an effective communication tool among experts \cite{Steindl.2021}. In addition to describing maintenance activities in the form of tasks, events and gateways can also be included\cite{Rosing.}. Service tasks are suitable for integrating web services into the maintenance process. As such, they are placeholders for digital services to be built. Manual or user tasks, performed by the technicians, can also be combined with digital services. This is particularly relevant in maintenance, where human intervention cannot be fully replaced by automation. If necessary, data flows and necessary data sources can be represented. By combining \textit{User Stories }and the \textit{BPMN process model}, a comprehensive understanding of the information needs and the maintenance process is created.

In \textit{Step 2}, it is necessary to specify the ontology requirements. Based on the \textit{User Stories} and the \textit{BPMN process model}, software experts can detail the software architecture. Here, individual components of the architecture are modeled using UML. A critical part of this design involves modeling the architecture's components and defining interfaces where ontology requirements are articulated in the form of \textit{CQs}. These are documented in an \textit{Ontology Requirements Specification Document (ORSD)} and needed as input in the following ontology building steps. This documentation captures both the \textit{CQs} and their anticipated \textit{answers}. At this juncture, categorizing the \textit{CQs }enhances clarity and focus. First, \textit{CQs} relying on existing knowledge, such as engineering information necessary for the second phase, are noted. Second, \textit{CQs} that require answers from the developed \textit{data-driven model} are specified. These are essential for the third phase, pending successful evaluation (\textit{Step 8}) of models based on operational and maintenance data. Domain and software experts take on a developing role in this step. Their collaboration is essential, with software experts contributing their knowledge of data sources, architectures, and technologies vital for the assistance application. Meanwhile, domain experts provide in-depth insights into CPS and corrective maintenance activities. Their joint effort aims to get a unified vision of the application to be developed.

\subsection{Contextualisation of CPS Life Cycle Data}
In \textit{Step 3}, an application-specific \textit{LWO (UML class diagram) }is built, guided by the \textit{CQs} from the \textit{ORSD }identified earlier. Here, domain and software experts engage in development roles again. This ensures adherence to the domain expert-centric approach. Domain experts bring their understanding of the necessary concepts and relationships for modeling, pinpointing critical information for CPS maintenance. These can be used accordingly by data experts to identify relevant features for modeling. Meanwhile, software experts translate this domain knowledge into the \textit{LWO}. To streamline this process, leveraging existing modular, standards-based, domain-specific \textit{lightweight ODPs }is recommended, ensuring consistent semantics. The linkage of individual \textit{ODPs} is employed by using four ontology alignment mechanisms (equivalent-to, subclass, attribute-to-class, relation-to), facilitating tailored relationships between concepts for the targeted application \cite{Hildebrandt.2020}. In particular, the subclassing option is suitable for achieving the necessary level of abstraction from a domain expert's perspective. Knowledge pertinent to the CPS, encompassing its structure, function, behavior, and fault diagnosis holds significant importance for corrective maintenance \cite{Gill.2023b, Steindl.2021, Gill.2023a}. Thus, the provision of ODPs enriched with such knowledge promises high reusability. Numerous \textit{light-} and \textit{heavyweight ODPs}, demonstrating extensive reusability across various CPS life cycle use cases, have been previously developed by the authors \cite{Gill.2023a, Gill.2023b, Hildebrandt.2020}.  A selection of the mentioned \textit{ODPs} is publicly available\footnote{VDI 2206, VDI 3682, DINEN61360: https://github.com/hsu-aut/}. In this application case for CPS maintenance, the \textit{ODPs VDI 3682} and \textit{VDI 2206} can be effectively reused to describe the functional and structural aspects of the CPS \cite{Hildebrandt.2020, Gill.2023b}. Additionally, the \textit{Semantic Sensor Network Ontology (SSN/SOSA) }is often helpful for detailing sensor and actuator data crucial for behavioral model development. Moreover, the \textit{ODP DIN EN 61360} facilitates semantic property descriptions of CPS. The \textit{ODP ISO 17359 }supports fault diagnosis and condition monitoring by considering relationships between faults, fault symptoms, and diagnostic models. For emerging knowledge needs, other information resources, especially domain-specific standards, may be analysed to create new, reusable \textit{lightweight ODPs} for future applications. 

The \textit{LWO} developed serves as a foundation for ontology experts to build an application-specific \textit{HWO} with \textit{Protege} (\textit{Step 4}), incorporating existing \textit{ODPs}. Currently, methods towards automating this process exist. For example, the \textit{Chowlk Converter} facilitates the automatic generation of a \textit{HWO} in OWL format from a UML class diagram \cite{ChavezFeria.2022}. Libraries necessary for this conversion can be provided for the \textit{LWO} building step. Software experts can utilize tools like \textit{draw.io} for \textit{LWO} modeling. Subsequently, ontology experts are tasked with formulating the requisite \textit{SPARQL queries} (e.g., \texttt{SELECT}) or updates (e.g., \texttt{INSERT}), in order to answer the \textit{CQs}. Following this, \textit{mappings (e.g. RML and R2RML)} essential for populating the \textit{HWO} are established. Additionally, the newly formulated and verified \textit{ODPs} are published (e.g. with a W3ID) and archived for subsequent applications. To aid in data preparation, employing inference or consistency rules (via \textit{SWRL} or \textit{SHACL}) respectively integrating a reasoner proves beneficial in detecting and addressing data inconsistencies.

During the data understanding (\textit{Step 5}) and preparation (\textit{Step 6}) steps, previously created ontological artifacts are put to use. Data scientists and analysts draw on these resources, particularly the \textit{LWO}, to deepen their domain knowledge as well as access and analyze existing data. This base enables easier identification of key features for model training. Access to necessary data is facilitated via \textit{SPARQL} endpoints, allowing for analysis and hypothesis formation. In the data preparation step, data engineers employ \textit{SPARQL} queries alongside \textit{inference} and \textit{consistency rules} to establish a data pipeline focused on processing data for feature extraction. As noted earlier, the use of informed ML is becoming increasingly prevalent. Consequently, the ontology can be transformed into knowledge graph embeddings, a vital step for subsequent ML training \cite{Rueden.2021, Sahlab.2021}. These embeddings provide a numerical representation that can be efficiently processed by ML algorithms, facilitating the detection of complex patterns vital for maintenance tasks like fault diagnosis. The modeling step (\textit{Step 7}) then utilizes this data to derive new insights through \textit{data-driven models}.

\subsection{Semantic Annotation of Extracted Information}
Initial steps of the third phase involve rigorously evaluating the \textit{data-driven model} (\textit{Step 8}) using diverse metrics. This evaluation by data experts is crucial within the CRISP-DM, significantly influencing the project's success. The domain expert's consulting role is essential, in order to assess the accuracy of model results. If the \textit{data-driven model} (\textit{DM} or \textit{ML}) fulfills predefined criteria, semantic annotation and ontology alignment steps follow. This involves revisiting and expanding the T-Box with additional concepts and relations, marked by the semantic annotation of the model's outputs. 

The \textit{LWO} expansion (\textit{Step 9}) requires collaboration between domain and data experts to define relevant concepts and relations. Apart from the developed \textit{data-driven model}, this step also utilizes ontological artifacts that were created earlier (\textit{LWO, ORSD}) as well as new ones (existing \textit{LWO Design Patterns, relevant information sources}). The incorporation of specific concepts and relations varies depending on the insights from the \textit{data-driven model}. Again, tools like the \textit{Chowlk converter} and \textit{draw.io}, as mentioned above, can automate the \textit{HWO} expansion in this context. Additionally, the description of behavioral information can be updated using the newly developed \textit{data-driven model }(e.g. regression model). This ensures that outdated system information does not compromise corrective maintenance activities. 

In the \textit{HWO} expansion step (\textit{Step 10}), all necessary ontological artifacts, especially the extended \textit{HWO} as well as necessary \textit{rules} are developed by the ontology expert, leveraging the existing artifacts (\textit{ORSD}, extended \textit{LWO}, existing \textit{HWO},\textit{ rules} and \textit{ODPs}). This step may also establish \textit{mappings} between \textit{data-driven model} outputs and ontology instances. This is especially relevant for fault diagnosis, where identified fault symptoms need to be combined with fault classes. Another option is to define \textit{SPARQL} \texttt{INSERT} requests to update the ontology with model outputs. The deployment step (\textit{Step 11}) enables software experts to utilize the \textit{data-driven model} and ontological artifacts (\textit{extended HWO} and \textit{rules}) to develop new digital services for assistance applications. The \textit{BPMN process model} created in \textit{Step 1} can be reused to orchestrate the developed digital services according to the previously defined maintenance process using a workflow engine \cite{Steindl.2021}.

\section{Case Study}\label{Evaluation}
\begin{figure}[ht]
\centering
\includegraphics[width=0.42
\textwidth]{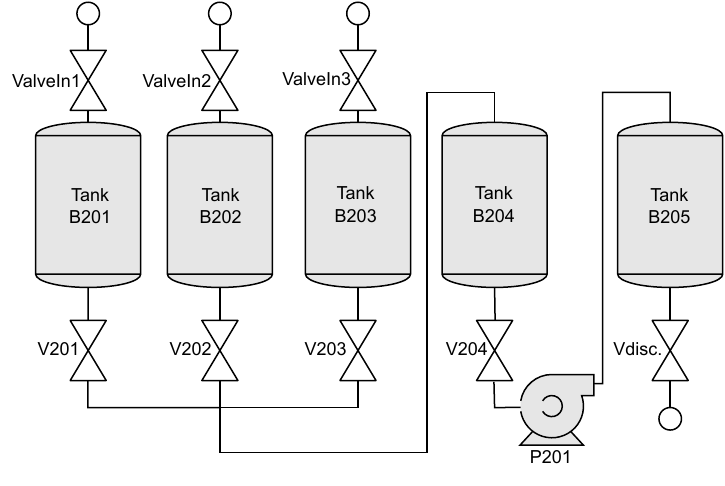}
\caption{Mixing module of the process plant}
\label{fig:MixingModule}
\end{figure}

The practical application of the method outlined in Sec. \ref{contribution} is demonstrated by an anomaly detection task within a modular process plant. The mixing module analyzed includes five tanks, as depicted in Fig. \ref{fig:MixingModule}. The operational process involves transferring liquids from three tanks (B201-B203) to a mixing reservoir (B204), and then pumping the mixture into tank B205 via pump P201. After emptying tank B205, the cycle repeats. Monitoring is achieved through various sensors that measure tank levels, temperatures, and flow rates, with all data being captured in a database. Actuator states are also stored, providing context information about process control. Additional information about the structure and function of the plant was obtained from the P\&I diagram, as well as existing CSV and JSON files. To evaluate the method, the high-fidelity simulation model provided by \cite{Ehrhardt.2022} was used. This model allows for the simulation of leakages and pipe blockages. The deployment of digital services for this process was facilitated by the semantic microservice framework introduced by Steindl et al. \cite{Steindl.2021}, which aligns with the Reference Architectural Model Industry 4.0 (RAMI 4.0) layers. Within this framework, the business layer includes the maintenance workflow using BPMN 2.0, which is automated by a workflow engine (Zeebe) in the functional layer. This layer also orchestrates the individual digital services, with inter-service communication handled by a message-oriented middleware (Apache Kafka). The developed application-specific ontology, residing in the information layer, both informs the digital services and gets updates with new data from these services.

\subsection{Ontology Requirements Specification}

\begin{figure}[t]
\includegraphics[width=0.48\textwidth]{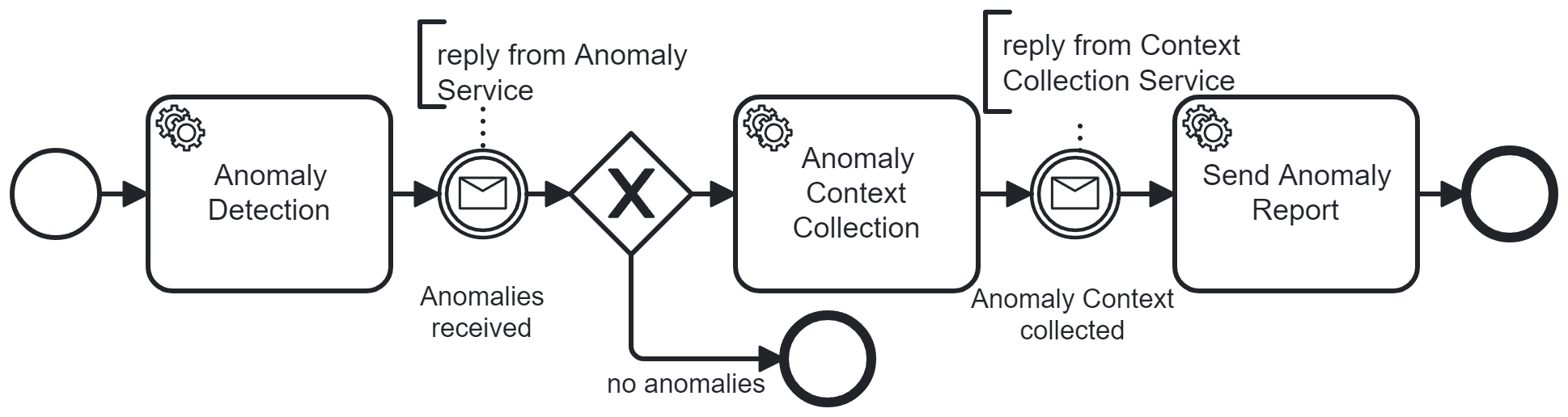}
\caption{Anomaly detection process modeled with BPMN}
\label{fig:BPMN}
\end{figure}

As outlined in Sec. \ref{contribution}, the initial steps involved business understanding and ontology requirements specification (\textit{Step 1} and \textit{Step 2}). In \textit{Step 1}, \textit{User Stories} were filled out using predefined templates. A key concern for the domain experts was the reliable detection of anomalies in the system in order to implement effective corrective maintenance activities. Given the considerable effort required to manually update the behavior model for anomaly detection, it was decided to learn this model from collected data. Moreover, another problem in the past was the interpretation of detected anomalies due to a lack of prior knowledge. This prior knowledge, dispersed across various artifacts from the engineering phase, had not been incorporated into the analysis. Therefore, upon identifying anomalies, a report should be generated that not only detailed the anomalies but also provided additional context information about the mixing module. Additionally, the relevant segment of the maintenance process, specifically concentrating on anomaly detection and analysis, was modeled using\textit{ BPMN }(see Fig. \ref{fig:BPMN}). Based on the preceding \textit{User Stories}, three new digital services (modeled as service tasks) were required. Firstly, an \textit{Anomaly Service} was needed to detect anomalies based on operational data as well as the learned behavioral model. Secondly, a \textit{Context Collection Service} was essential to provide additional contextual information and prior knowledge about the detected anomalies based on further system information. Subsequently, all the collected information should be consolidated by the \textit{Report Service} to send a report to technicians, enabling them to determine possible fault causes. Besides, the project requirements defined general objectives and criteria, such as quality and termination conditions. In \textit{Step 2}, two types of \textit{CQs} were developed along with their answers and documented in the \textit{ORSD}. Firstly, \textit{CQs} were formulated that contributed to the contextualization of CPS data from the life cycle. These included questions about function (\textit{e.g., "Which part of the module is responsible for filling tank B201?"}), structure (\textit{e.g., "Which sensors are part of tank B201?"}), and behavior (\textit{e.g., "What property does the sensor at tank B201 measure?"}). This information was intended to assist data experts in understanding the data and its preparation. Furthermore, this information was partially needed for the \textit{Context Collection Service}. Secondly, \textit{CQs} were defined that directly related to the original goal of anomaly detection. These included \textit{CQs} like \textit{"Between which two states was a temporal anomaly identified?"} or \textit{"By how many seconds was the anomaly outside the max?"}. The answers to these questions could be provided with the help of the learned model as well as identified timing anomalies (see phase three). 

\subsection{Contextualisation of Mixing Module Life Cycle Data}
\begin{figure*}[ht]
\centering
\includegraphics[width=0.94\textwidth]{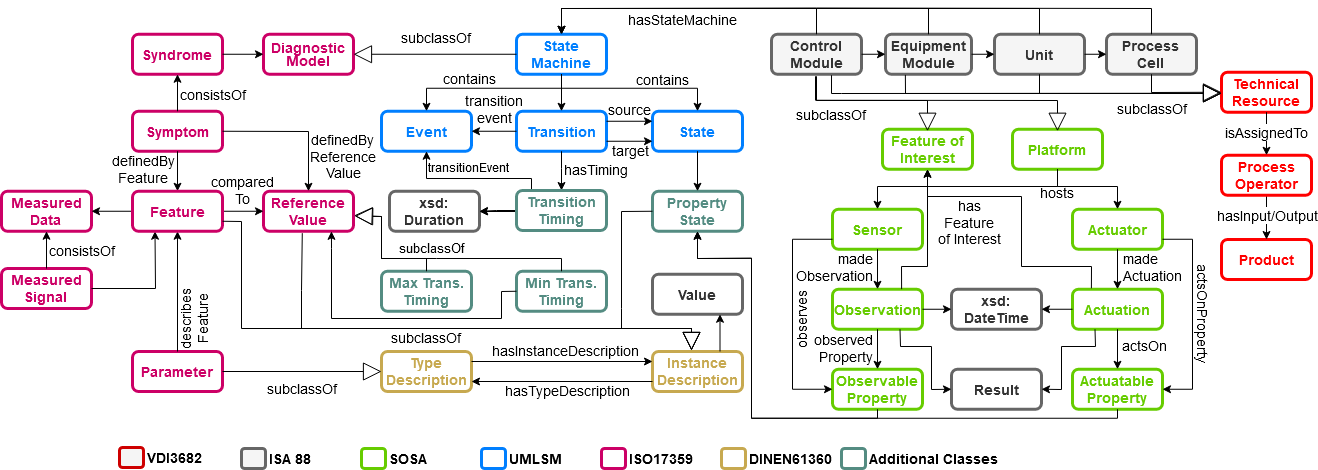}
\caption{Excerpt from the LWO for the case study}
\label{fig:LWO}
\end{figure*}
To aid data experts with information from the life cycle of the mixing module, an application-specific \textit{HWO} was developed in phase two. Software and domain experts drew upon existing domain-specific \textit{ODPs} in \textit{Step 3}. The \textit{ODP VDI 3682} was utilized for the functional description of the CPS. For structure information, \textit{ODP ISA 88} was applied, offering detailed descriptions of plant hierarchy and process recipes. Additionally, the \textit{SOSA ontology} was utilized to link sensor and actuator information with behavioral data. The\textit{ ODP DIN EN 61360} was employed for property description of the system. By aligning these \textit{ODPs}, an application-specific \textit{LWO} of the mixing module was created. During \textit{Step 4}, an ontology expert created the HWO. Based on the \textit{LWO} and the existing \textit{ODPs}, all classes as well as object and data properties were modeled using \textit{Protege}. Mappings from the CSV and JSON files, as well as the database, were performed using \textit{RML} and \textit{R2RML} to integrate both static and dynamic data. The OBDA tool \textit{Ontop} \cite{Calvanese.2016} facilitated virtual access to individual sensor and actuator data from the relational database. Created \textit{SPARQL} queries provided data experts with access to the ontology. Additionally, the \textit{LWO} was made available to data experts in \textit{Step 5}. The most extensive work of data integration in \textit{Step 6} for data preparation was accomplished through the development of the \textit{HWO}. In \textit{Step 7}, a python implementation of the \textit{OTALA algorithm} was selected to learn a \textit{Timed Automata} that depicts the module's behavior \cite{A.Maier.2015}. This learned model features one initial state and six production states, each incorporating actuator information and timing distributions. The \textit{Anomaly Detection algorithm ANODA} was used for anomaly identification \cite{A.Maier.2015}. For this purpose, the fault cases described at the beginning were simulated to collect an appropriate test set. This set, alongside the \textit{Timed Automata}, was then passed to the algorithm to identify anomalous behavior. A detailed description of this approach, including the algorithms used, can be found in \cite{Westermann.2023}.

\subsection{Semantic Annotation of Extracted Timed Automata and Anomalies}
In the third phase of the method, the focus was on the semantic annotation of the newly extracted information (see \textit{Step 8}). The goal was to answer the \textit{CQs} from the \textit{ORSD} defined in phase one regarding the anomaly detection activity. Consequently, it was necessary to align the information generated through the CRISP-DM, especially the \textit{Timed Automata} and identified anomalies, with the existing ontology. During \textit{Step 9}, the domain and data experts expanded the existing \textit{LWO}. The \textit{Timed Automata} was described using the \textit{ODP UML State Machine}, which outlines the system's states $S$, transitions $T$, and events $\Sigma$. The \textit{ODP ISO 17359}, already applied in other contexts for fault diagnosis \cite{Gill.2023a}, was also utilized for describing anomalies. In this particular case, the state machine was modeled as a subclass of the diagnostic model, which detects timing anomalies (defined as symptoms in \textit{ISO 17359}). It was also aligned with the structural information of \textit{ODP ISA 88}. Linking anomalies to the clearly defined states and events of the state machine simplified the interpretation of anomalies and facilitated the identification of suitable countermeasures. An excerpt of the expanded \textit{LWO} with its alignments is shown in Fig. \ref{fig:LWO}. In the subsequent \textit{Step 10}, the expanded \textit{HWO} was built and \textit{mappings} were conducted to detail the A-Box. The relevant ontological artifacts, including the complete expanded \textit{HWO}, defined \textit{SPARQL} requests, \textit{rules (e.g. mappings)} as well as the \textit{BPMN process model} were made available to software experts in \textit{Step 11}. With the help of the UML models and the developed digital services, the application-specific \textit{HWO} could finally be integrated into the semantic microservice framework for the assistance application. The \textit{Timed Automata} and the \textit{ANODA algorithm} were implemented in the \textit{Anomaly Service}, which accessed the operational data via \textit{OBDA}. The \textit{SPARQL} queries for additional context regarding the identified anomalies were included in the \textit{Context Collection Service} to subsequently generate and send the report to the technician.

\section{Discussion}\label{Discussion}
In the following, the method will be discussed based on two criteria: Firstly, the assessment will determine whether required knowledge has been successfully represented. Secondly, the method will be analysed in terms of its efficiency.

Two key observations emerge with regard to the first discussion criterion: Firstly, all \textit{CQs} in the case study were successfully answered. Some \textit{CQs} helped contextualize CPS life cycle data for data experts using the alignment of \textit{ODPs VDI 3682}, \textit{ISA88}, \textit{SOSA}, and \textit{DIN EN 61360}. Others were addressed through data-driven extraction of the \textit{Timed Automata} as well as anomalies following the CRISP-DM. Using these artifacts in combination with \textit{ODP ISO 17359} and the \textit{UML State Machine} the existing ontology could be populated (A-Box) as well as expanded (T-Box). Overall, the targeted digital services were successfully developed, allowing for the collection of relevant information through their combination. A detailed enumeration of the \textit{CQs} and \textit{answers} used can be found in \cite{Westermann.2023}. Secondly, integrating engineering knowledge with data-driven insights from operational and maintenance phases was essential to develop the application-specific ontology. All steps of the method, in the defined sequence, were relevant in this context (phases 1-3, see \textbf{R1-R3}).

Regarding the second discussion criterion, the use of reusable \textit{ODPs} was a key factor to enhance efficiency (see \textbf{R4}). The development of the application-specific ontology was accelerated by reusing \textit{ODPs}, previously applied in various publications by the authors (\cite{Gill.2023a, Gill.2023b, Hildebrandt.2020}).  While the \textit{ODPs SOSA} and \textit{UML State Machine} were not developed by the authors, they were reused for this case study. The method presented was also applied to an aerospace maintenance use case (see \cite{Gill.2023b}), focusing on a completely different CPS (aircraft component), another data-driven model as well as further maintenance activities (e.g. testing, fault diagnosis, maintenance planning). Many ontological artifacts created for this aircraft maintenance use case, which involved data-driven fault classification based on test bench and life cycle data, could be reused. In addition, the involvement of various experts and the provision of comprehensible ontological artifacts, such as the \textit{LWO}, enabled efficient collaboration (see \textbf{R5}). However, improving the efficiency of ontology population is necessary, as modeling A-Box instances and defining mappings remain labor-intensive.

\section{Summary And Outlook}\label{Summary and Outlook}

This contribution introduced a method for the efficient development of an application-specific ontology for corrective maintenance of CPS. In this context, the CRISP-DM was combined with a domain-expert-centric approach to ontology design. Firstly, based on a systematic specification of ontology requirements, ontological artifacts for contextualizing CPS lifecycle data were integrated as input into the steps of data understanding and data preparation. Secondly, data-driven models developed through the CRISP-DM were semantically annotated and aligned with the existing ontology. Based on the created ontological artifacts (e.g., HWO, LWO), digital services for assistance applications could subsequently retrieve from or update information in the ontology. The evaluation of the method was conducted using a case study. In this case, an ontology was developed to describe the function, structure, behavior as well as diagnostic information of a modular mixing plant. With the help of a modular domain ontology, a Timed Automata was learned, and anomalies were identified. Using ODPs such as the UML State Machine and ISO 17359, this information could be used to expand the existing ontology.

With view to future work, there are additional hurdles to overcome and improvement opportunities to exploit. Particular caution is warranted when making inferences based on the expanded ontology. Especially in fault diagnosis, when integrating data-driven models or incomplete data, considering uncertainty is essential. Besides, methods and tools to utilize engineering or runtime artifacts for mapping are needed, since instantiating the expanded ontology still requires substantial effort. However, the combination approach introduced also offers promising prospects. The method is suitable for Informed ML approaches within fault diagnosis. Specifically, incorporating domain knowledge through knowledge graph embeddings can decrease the amount of data required for training. Thus, the detection of complex patterns and relationships critical for fault diagnosis is facilitated, even with limited data. Beyond corrective maintenance, the integration of formalized domain knowledge with data-driven approaches is also suitable for predictive and prescriptive maintenance. Regardless of the maintenance application, the third phase can be utilized to iteratively refine the initially developed ontology using a data-driven approach. The modular structure allows for efficient updates based on operational insights without the need to revise the entire ontology.

\section*{Acknowledgment}
This research [project ProMoDi] is funded by dtec.bw – Digitalization and Technology Research Center of the Bundeswehr. dtec.bw is funded by the European Union – NextGenerationEU.

\bibliographystyle{IEEEtran}

\vspace{12pt}
\end{document}